%% file: seminovo.tex
\documentclass[sigconf,balance=false]{acmart}

\AtBeginDocument{}
\setcopyright{none}
\renewcommand\footnotetextcopyrightpermission[1]{}
\acmYear{2027}
\acmDOI{}
\acmConference[KDD '27]{KDD 2027 AI for Sciences Track}{August 1--5, 2027}{San Jose, CA, USA}
\usepackage{booktabs}
\usepackage{multirow}
\usepackage{siunitx}
\usepackage{xcolor}
\usepackage{amsmath}
\usepackage{graphicx}
\usepackage{microtype}
\usepackage{placeins}
\usepackage{algorithm}
\usepackage{algpseudocode}
\usepackage{enumitem}

\graphicspath{{figures/}}
\definecolor{seminovoblue}{HTML}{2474A6}
\definecolor{seminovogreen}{HTML}{2A9D8F}
\definecolor{seminovoorange}{HTML}{E76F51}
\newcommand{\method}{\textsc{SemiNovo}}
\newcommand{\dataset}{\textsc{DarkSpec}}
\newcommand{\denovo}{\textit{de novo}}
\newcommand{\DeNovo}{\textit{De Novo}}

\begin{document}

\title{SemiNovo: Learning Beyond Search-Identified Spectra for \DeNovo{} Peptide Sequencing}

\author{Pan Liu}
\authornote{These authors contributed equally to this work.}
\email{liup28292@gmail.com}
\affiliation{
  \institution{Information Hub, HKUST(GZ)}
  \city{Guangzhou}
  \country{China}
}

\author{Dongxin Lyu}
\authornotemark[1]
\email{lyudongxin@westlake.edu.cn}
\affiliation{
  \institution{Institute of Engineering, Westlake University}
  \city{Hangzhou}
  \country{China}
}

\author{Yusen Tan}
\email{ytan277@connect.hkust-gz.edu.cn}
\affiliation{
  \institution{Information Hub, HKUST(GZ)}
  \city{Guangzhou}
  \country{China}
}

\author{Yifan Li}
\email{yli994@connect.hkust-gz.edu.cn}
\affiliation{
  \institution{Information Hub, HKUST(GZ)}
  \city{Guangzhou}
  \country{China}
}

\author{Hai-tao Yu}
\email{hyu382@connect.hkust-gz.edu.cn}
\affiliation{
  \institution{Information Hub, HKUST(GZ)}
  \city{Guangzhou}
  \country{China}
}

\author{Yunhua Zhong}
\email{yunhuazhong@connect.hku.hk}
\affiliation{
  \institution{Department of Chemistry, The University of Hong Kong}
  \city{Hong Kong}
  \country{China}
}

\author{Zikun Wang}
\email{ericwanglzu@outlook.com}
\affiliation{
  \institution{Information Hub, HKUST(GZ)}
  \city{Guangzhou}
  \country{China}
}

\author{Tianze Ling}
\email{tianzeling98@outlook.com}
\affiliation{
  \institution{School of Life Sciences, Tsinghua University}
  \city{Beijing}
  \country{China}
}

\author{Jun Xia}
\authornote{Corresponding author.}
\email{junxia@hkust-gz.edu.cn}
\affiliation{
  \institution{Information Hub, HKUST(GZ)}
  \city{Guangzhou}
  \country{China}
}

\begin{abstract}
Sensitive peptide identification is fundamental to protein characterization and novel protein discovery, for which \denovo{} peptide sequencing (DNPS) from tandem mass spectra provides an important database-independent approach. Although deep learning has substantially advanced \denovo{} sequencing, existing models are predominantly trained on high-confidence peptide--spectrum matches obtained through database search. This search-conditioned supervision excludes a large fraction of unidentified spectra, creating a distribution mismatch between the selectively labeled training data and the complete spectrum stream encountered at inference time, thereby limiting peptide identification performance. To address this challenge, we propose \method{}, a semi-supervised framework that learns jointly from identified and unlabeled spectra. \method{} first constructs \dataset{}, a large-scale quality-controlled corpus of unlabeled spectra collected from diverse public proteomics projects, thereby extending training beyond search-identified spectra and mitigating the resulting shift in DNPS supervision. It then combines stepwise mass-constrained autoregressive decoding with cumulative path confidence to derive reliable supervision from unlabeled spectra. Moreover, \method{} introduces head-wise chemical-evidence gating to improve spectral evidence aggregation under noisy and incomplete fragmentation. Experiments show that \method{} performs favorably against state-of-the-art methods on multiple benchmarks. \href{https://github.com/grandOrgan/Seminovo}{Code} and \href{https://huggingface.co/datasets/PanLiu/DarkSpec}{dataset} are publicly available.
\end{abstract}



\maketitle

\noindent\textit{Accepted as an oral presentation at KDD 2027, AI for Sciences Track.}

\input{sections/00_teaser}
\input{sections/01_introduction}
\input{sections/02_related_work}
\input{sections/03_method}
\input{sections/04_experiments}
\input{sections/05_conclusion}
\input{sections/06_limitations_ethics}
\input{sections/07_genai_usage}

\clearpage
\setlength{\bibsep}{1.5pt plus 0.2pt}
\bibliographystyle{ACM-Reference-Format}
\bibliography{references}

\clearpage
\input{sections/06_appendix}

\end{document}

%% file: sections/00_teaser.tex
\begin{figure}[!t]
  \centering
  \includegraphics[width=\columnwidth]{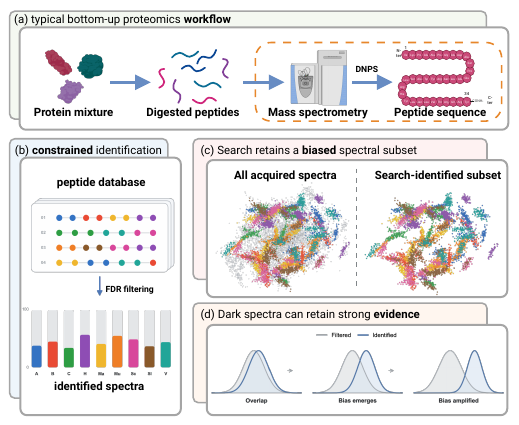}
  \caption{\textbf{Database search induces biased supervision.} (a) Tandem mass spectrometry fragments peptides into spectra that must be assigned amino-acid sequences. (b) Candidate-database search and false discovery rate control identify only part of the acquired spectra. (c) Retained peptide--spectrum matches occupy a selective subset of the complete spectral space. (d) Increasing identification stringency separates retained and excluded distributions, leaving informative dark spectra outside training.}
  \Description{Four panels introduce tandem mass spectrometry peptide identification and illustrate search-conditioned supervision bias through database filtering, selective coverage of spectral representation space, and the growing distribution separation between identified and dark spectra.}
  \label{fig:motivation}
\end{figure}

%% file: sections/01_introduction.tex
\section{Introduction}
\label{sec:introduction}

Identifying proteins in complex biological samples is fundamental to understanding cellular processes, elucidating disease mechanisms, and discovering biomarkers and therapeutic targets~\cite{aebersold2016proteome,geyer2017biomarker}. Tandem mass spectrometry (MS/MS) is the principal high-throughput technology for large-scale protein identification~\cite{aebersold2003proteomics}. As illustrated in Figure~\ref{fig:motivation}(a), proteins are digested into peptides, precursor ions are isolated and fragmented, and the mass-to-charge ratios and intensities of the resulting fragment ions form an MS/MS spectrum. Because cleavage along a peptide backbone produces mass differences associated with amino-acid residues, the spectrum provides a chemical fingerprint from which the underlying sequence can be inferred~\cite{zhang2013shotgun}. Database search assigns this fingerprint to a peptide in a predefined candidate collection. By contrast, \denovo{} peptide sequencing (DNPS) directly predicts the sequence from the spectrum and precursor information, without requiring the correct peptide to occur in a database. This property makes DNPS valuable for identifying sequence variants, unexpected post-translational modifications (PTMs), and peptides beyond reference proteomes~\cite{tran2017deepnovo,yilmaz2022casanovo}.

Deep learning has substantially advanced DNPS, progressing from recurrent models over discretized spectra to point-based encoders, Transformers, retrieval-augmented generation, and diffusion models~\cite{tran2017deepnovo,qiao2021pointnovo,yilmaz2022casanovo,yilmaz2024casanovo,liu2023pepnet,yang2024helixnovo,eloff2023instanovo,zhang2025primenovo,chen2024renovo,chen2026diffunovo}. Yet these architectures share a dependence on database-search supervision~\cite{yilmaz2022casanovo,zhou2024novobench}. Search engines score candidate peptide--spectrum matches (PSMs)~\cite{cox2008maxquant,kim2014msgf,kong2017msfragger,yu2025deepsearch,a2026ddabert}, and only PSMs passing target--decoy false discovery rate (FDR) control are retained as labels~\cite{levitsky2017fdr}. Spectra remain unlabeled when the peptide is absent from the database, fragmentation evidence is incomplete, or the match score is insufficient. Many such spectra recur across experiments and collectively form a substantial, structured population~\cite{griss2016unidentified}. We refer to them as dark spectra. Figure~\ref{fig:motivation}(b--d) shows how candidate restriction and confidence filtering preserve only a selective region of the acquired spectral space. Consequently, a model that is database-independent at inference remains trained on the subset that database search can identify.

Learning from dark spectra is therefore important, but it presents three coupled challenges. First, identified PSMs favor database-covered peptides and interpretable fragmentation, whereas dark spectra exhibit different peak density, noise, and fragment completeness. This domain shift encourages models to rely on clean ion ladders and familiar peptide compositions that may not transfer to the full acquisition stream~\cite{griss2016unidentified,xia2024adanovo}. Second, dark spectra provide no peptide annotations. Pseudo-labeling is possible, but autoregressive sequencing amplifies an early residue error through every later decoder state and prefix-mass decision; token-wise confidence alone cannot ensure that the resulting path remains reliable~\cite{sohn2020fixmatch,liu2025confidence,patel2023sequps}. Third, dark spectra often contain weak, noisy, or incomplete fragmentation evidence~\cite{mao2023graphnovo}. Our analysis further finds that decoder attention can concentrate on structurally privileged prefix positions rather than residue context supported by diagnostic fragments. This misplaced concentration wastes representational capacity precisely when chemically informative evidence is scarce.

We propose \method{}, a unified framework comprising an acquisition-derived dataset, a chemical-evidence-gated model, and a chemically structured semi-supervised training paradigm. First, \dataset{} is refined from 44.9 million spectra collected across public projects. Quality filtering removes malformed and fragment-poor observations, while near-duplicate reduction prevents recurring acquisitions from dominating training; 4.5 million unlabeled spectra broaden coverage beyond search-identified PSMs. Second, \method{} generates pseudo-labels through stepwise mass-constrained autoregressive decoding. Cumulative path confidence retains the longest reliable prefix and terminates generation once the path becomes unreliable, preventing an early error from contaminating later supervision. Third, query-dependent chemical-evidence gates regulate how strongly attention output enters the residual stream. This mechanism suppresses updates dominated by structurally special but chemically weak positions and preserves sequence-informative fragment evidence under ambiguous fragmentation. Experiments show that \method{} performs favorably against state-of-the-art methods on multiple benchmarks. Our contributions are summarized as follows.
\begin{itemize}[leftmargin=1.35em,labelsep=0.45em]
  \item \textbf{Dataset.} We characterize the impact of search-conditioned distribution shift on DNPS and construct \dataset{}, a broad-source corpus of 4.5 million unlabeled spectra refined through quality filtering and near-duplicate clustering.
  \item \textbf{Semi-Supervised Learning.} We develop a DNPS-specific strategy based on stepwise mass-constrained decoding, cumulative path confidence, reliable-prefix supervision, and monotone early termination.
  \item \textbf{Model.} We identify systematic attention concentration on chemically weak prefix positions and introduce head-wise chemical-evidence gating to improve the use of sparse and incomplete fragment evidence.
  \item \textbf{Performance.} \method{} achieves state-of-the-art peptide precision on multiple benchmarks and consistently outperforms its supervised counterpart across all three datasets.
\end{itemize}

\begin{figure*}[t]
  \centering
  \includegraphics[width=\textwidth]{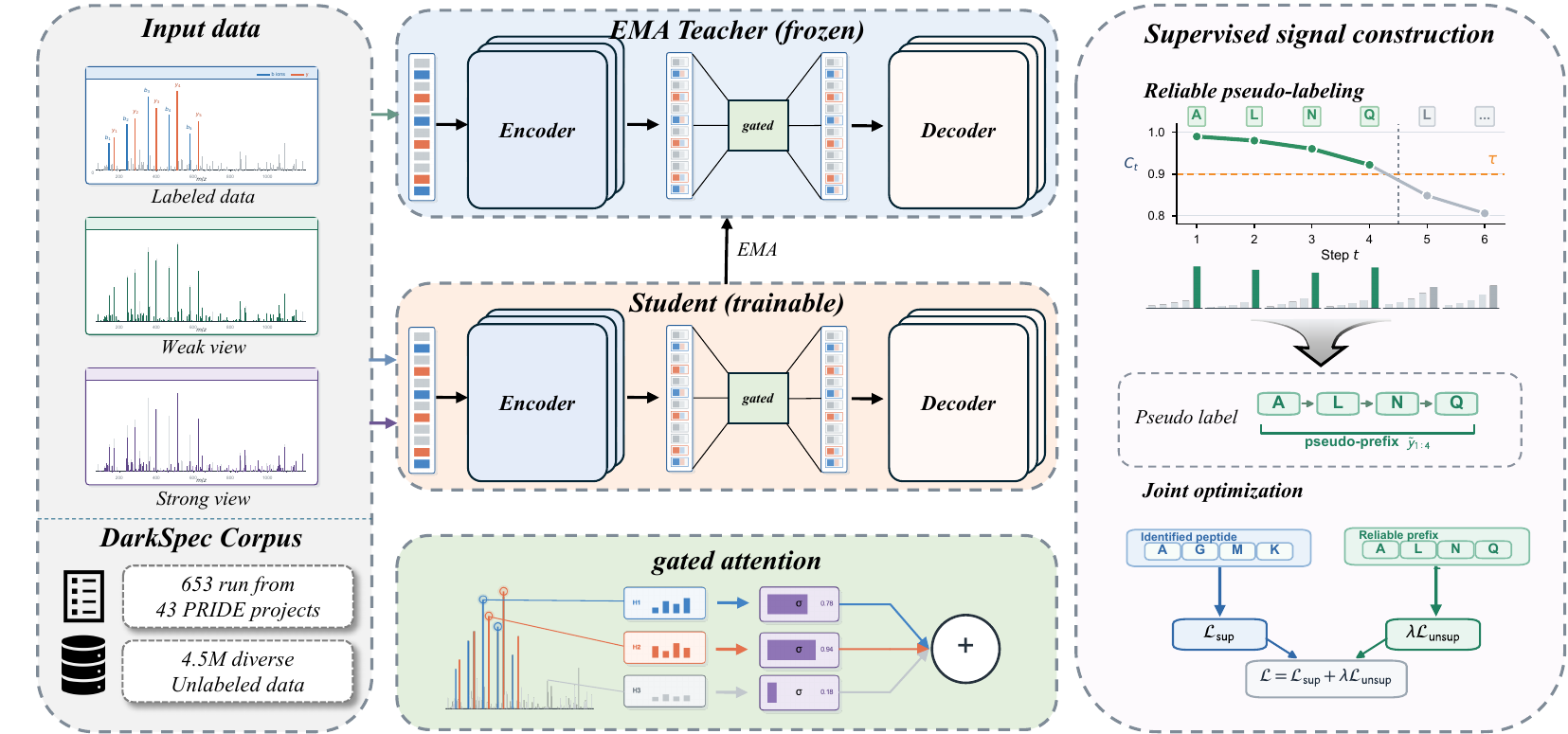}
  \caption{\textbf{Overview of \method{}.} Labeled, weak, and strong spectrum views enter vertically aligned exponential moving average (EMA) teacher and student networks with head-wise chemical-evidence gating. Mass-constrained autoregressive decoding retains the longest prefix whose cumulative path confidence exceeds $\tau$. The student jointly optimizes identified peptides and reliable pseudo-prefixes.}
  \Description{The figure shows labeled, weak, and strong spectrum views on the left; vertically aligned EMA teacher and student encoder--decoder networks with chemical-evidence gates in the center; and reliable pseudo-label construction and joint supervised and unsupervised learning on the right.}
  \label{fig:method}
\end{figure*}

%% file: sections/02_related_work.tex
\section{Related Work}
\label{sec:related}

\noindent\textbf{\DeNovo{} Peptide Sequencing.}
Classical \denovo{} peptide sequencing searches for residue paths that explain observed fragment ions using spectrum graphs or probabilistic models~\cite{frank2005pepnovo}. Deep learning replaces hand-crafted scoring with learned spectrum--sequence mappings. DeepNovo combines convolutional spectrum features with recurrent decoding~\cite{tran2017deepnovo}, PointNovo operates directly on unordered peaks~\cite{qiao2021pointnovo}, and Casanovo formulates sequencing as autoregressive Transformer generation~\cite{yilmaz2022casanovo}. Methods improve complementary aspects of this paradigm. AdaNovo addresses biases in PSMs~\cite{xia2024adanovo}, SearchNovo connects database search with \denovo{} sequencing~\cite{xia2024searchnovo}, ReNovo retrieves related peptides before generation~\cite{chen2024renovo}, and DiffuNovo introduces diffusion-based generation with explicit precursor-mass guidance~\cite{chen2026diffunovo}. NovoBench standardizes datasets, modified-sequence vocabularies, and evaluation metrics for these models~\cite{zhou2024novobench}. Nevertheless, both existing methods and benchmarks learn predominantly from confident database-search identifications. \method{} instead expands supervision to search-unidentified spectra while retaining the NovoBench protocol.

\noindent\textbf{Semi-Supervised Learning.}
Modern semi-supervised learning (SSL) commonly combines teacher--student consistency with pseudo-label selection. Mean Teacher stabilizes targets through an exponential moving average of model weights~\cite{tarvainen2017mean}, while FixMatch transfers confident predictions from weakly to strongly augmented views~\cite{sohn2020fixmatch}. FlexMatch adapts selection thresholds to class-dependent learning progress~\cite{zhang2021flexmatch}, and USB shows that SSL behavior varies substantially across tasks and domains~\cite{wang2022usb}. Confidence is therefore not automatically a reliable selection criterion. CSL further separates correct and incorrect predictions using their confidence distributions~\cite{liu2025confidence}. These studies primarily consider independent classification outputs. Peptide pseudo-labels are autoregressive and chemically constrained, so an early residue error changes every subsequent prediction. \method{} consequently evaluates cumulative path confidence, retains a contiguous reliable prefix, and restricts generation to residues compatible with the remaining precursor mass.

%% file: sections/03_method.tex
\section{Method}
\label{sec:method}

\subsection{Preliminaries}
\label{sec:preliminaries}
Let $x=\{(m_i,I_i)\}_{i=1}^{P}$ denote a tandem mass spectrum with $P$ peaks, where $m_i$ and $I_i$ are the mass-to-charge ratio and intensity of peak $i$. Its precursor is $c=(m_{\mathrm{pre}},z)$, comprising the precursor $m/z$ and charge. Given monoisotopic proton mass $m_{\mathrm{H}^{+}}$ and water mass $m_{\mathrm{H_2O}}$, the precursor defines the target residue-sum mass
\begin{equation}
 M(c)=z(m_{\mathrm{pre}}-m_{\mathrm{H}^{+}})-m_{\mathrm{H_2O}}.
 \label{eq:precursor-mass}
\end{equation}
A peptide is represented by $y=(y_1,\ldots,y_T)$, where $y_T=\mathrm{EOS}$. The vocabulary is $\mathcal V=\mathcal V_{\mathrm{res}}\cup\{\mathrm{EOS}\}$, where $\mathcal V_{\mathrm{res}}$ contains unmodified residues and each supported modified residue as an individual token. Let $\mu(a)$ be the residue mass of $a\in\mathcal V_{\mathrm{res}}$. The theoretical residue-sum mass is
\begin{equation}
 m(y)=\sum_{t:y_t\in\mathcal V_{\mathrm{res}}}\mu(y_t).
 \label{eq:peptide-mass}
\end{equation}
A valid peptide should satisfy the precursor-mass constraint $|m(y)-M(c)|\leq\delta$, where $\delta$ is the prescribed mass tolerance. Mass feasibility is evaluated at 50 ppm under isotope errors $\{0,1\}$, and EOS is permitted after at least six residues.

We consider a labeled set $\mathcal D_\ell$ of triples $(x,c,y)$ obtained from high-confidence PSMs and an unlabeled set $\mathcal D_u$ of pairs $(x,c)$ sampled from acquisition-derived source pools independently of database-search peptide assignments. The conditional peptide likelihood is factorized autoregressively
\begin{equation}
 p_\theta(y\mid x,c)=\prod_{t=1}^{T}p_\theta(y_t\mid y_{<t},x,c).
 \label{eq:ar-factorization}
\end{equation}
Here, $T$ includes the terminal $\mathrm{EOS}$ token. For a labeled minibatch $B_\ell\subset\mathcal D_\ell$, conventional supervised DNPS minimizes the teacher-forced token-average negative log-likelihood
\begin{equation}
 \mathcal L_{\mathrm{sup}}=
 -\frac{\displaystyle\sum_{(x,c,y)\in B_\ell}\sum_{t=1}^{T_y}
 \log p_\theta(y_t\mid y_{<t},x,c)}
 {\displaystyle\sum_{(x,c,y)\in B_\ell}T_y},
 \label{eq:sup}
\end{equation}
where $T_y$ is the number of non-padding targets in $y$. This objective learns only from search-identified spectra, leaving the much larger unlabeled spectrum population outside model optimization.

Figure~\ref{fig:method} presents \method{}, which complements supervised DNPS with three components. Canonical spectrum processing places labeled and unlabeled observations in a shared input space. A non-causal spectrum encoder and a chemically gated causal peptide decoder map spectra to residue sequences. Finally, stepwise mass-constrained autoregressive decoding converts unlabeled spectra into reliable pseudo-prefix supervision.

\subsection{DarkSpec Corpus}
\label{sec:corpus}
Supervised DNPS corpora retain spectra only when database search produces high-confidence PSMs. Their training distribution is consequently conditioned on database content and search sensitivity, while acquired spectra that remain unidentified receive no learning signal. To expose DNPS models to this omitted region, we construct \dataset{} from 653 acquisition runs across 43 public projects in the PRoteomics IDEntifications Database (PRIDE)~\cite{perezriverol2022pride}. Unlike a PSM corpus, inclusion depends on acquisition rather than peptide assignment. \dataset{} therefore mixes identifiable and unidentified spectra, approximating the acquisition-level population encountered at inference. Its source pool contains 44,963,379 MS/MS spectra from heterogeneous biological samples and experimental runs, covering spectra beyond conventional search-derived supervision. Appendix~\ref{app:darkspec} lists the contributing projects and spectrum counts.

The acquisition-derived pool broadens coverage but contains spectra with malformed precursors or insufficient fragment evidence. We therefore filter spectra by physical input quality. Using the same NovoBench-compatible processing for labeled and unlabeled data~\cite{zhou2024novobench}, we remove out-of-range, precursor-adjacent, and low-intensity peaks, retain the 150 most intense peaks, and normalize intensities. We discard malformed spectra, invalid precursor charges, and spectra with fewer than 20 valid peaks, retaining over 41 million spectra. Shared processing prevents an artificial domain gap between labeled and unlabeled inputs, while quality-based selection preserves the acquisition-derived unlabeled distribution.

Pooling many projects introduces a second bias: repeated exports and overlapping acquisitions can make a small number of recurring spectral patterns dominate the unlabeled objective. We group near-duplicate spectra using a fixed SimHash signature~\cite{charikar2002similarity,frank2008clustering,bittremieux2022embedding} derived from fragment patterns, precursor mass, and charge, and retain one representative per group. A deterministic subset of 4,500,000 representatives forms the final training corpus, controlling redundancy and computational cost. Exact filtering and redundancy-control settings are provided in Appendix~\ref{app:darkspec}.

\subsection{Chemical-Evidence Gating}
\label{sec:model}
\method{} uses a Transformer encoder--decoder for spectrum-to-peptide generation. Each peak is represented by independent multi-scale Fourier encodings of its continuous $m/z$ and normalized intensity~\cite{yilmaz2022casanovo,chen2026diffunovo}
\begin{equation}
\begin{aligned}
 \phi_{\Omega}(v)&=
 \big[\sin(\omega v),\cos(\omega v)\big]_{\omega\in\Omega},\\
 e_i&=F_{\mathrm{peak}}\!\left(
 [\phi_{\Omega_m}(m_i);\phi_{\Omega_I}(I_i)]\right),
\end{aligned}
 \label{eq:peak-embedding}
\end{equation}
where $\Omega_m$ and $\Omega_I$ are fixed frequency sets and $F_{\mathrm{peak}}$ is a two-layer projection to width $d$. A learned spectrum token is prepended, and a non-causal encoder models all peak representations bidirectionally because fragment peaks have no causal order. The decoder starts from a precursor query formed by the Fourier encoding of neutral precursor mass and a learned charge embedding; the encoded spectrum then conditions residue generation through cross-attention.

\begin{figure}[t]
  \centering
  \includegraphics[width=\columnwidth]{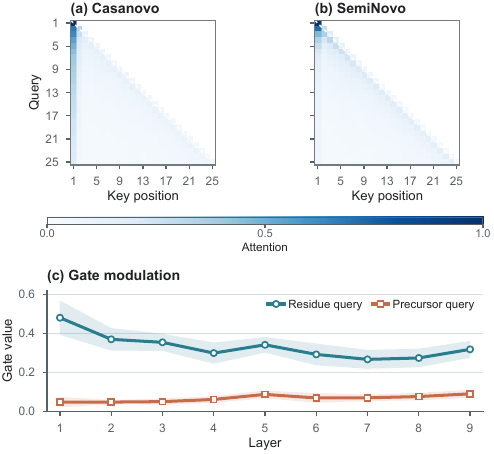}
  \caption{\textbf{Chemical-evidence gating suppresses non-localizing precursor-query updates.} (a--b) Mean decoder self-attention in Casanovo and SemiNovo. (c) Layer-wise SemiNovo gate values for precursor and residue-prefix queries. Smaller precursor-query gates reduce the influence of global precursor context on residue prediction. Lines and shaded regions denote mean and standard deviation.}
  \Description{Two decoder self-attention heatmaps compare Casanovo and SemiNovo, followed by a line plot showing that SemiNovo assigns consistently lower gate values to the precursor query than to residue-prefix queries across nine decoder layers.}
  \label{fig:attention-analysis}
\end{figure}

DNPS requires the decoder to distinguish residue-localizing fragment evidence from global precursor context. However, Transformer attention can over-emphasize structurally privileged tokens~\cite{xiao2024streamingllm,gu2024sink,sun2024massive,yu2024hidden}. As shown in Figure~\ref{fig:attention-analysis}, Casanovo allocates substantial attention to the precursor query while underweighting informative fragment evidence. This imbalance is especially harmful for dark spectra, whose informative ion series are often sparse or incomplete~\cite{zhang2013shotgun}. SemiNovo instead learns markedly smaller gates for the precursor query than for residue-prefix queries, attenuating chemically non-localizing updates before they enter the residual stream and preserving the decoder's capacity to integrate diagnostic fragment ions through cross-attention.

To regulate these prefix updates, \method{} applies chemical-evidence gating to causal self-attention in every peptide-decoder layer. For query states $H_q\in\mathbb R^{L\times d}$, an input-dependent sigmoid gate is computed for all $H$ attention heads~\cite{bondarenko2023quantizable,ramapuram2025sigmoid,qiu2025gated}
\begin{equation}
\begin{aligned}
 G&=\sigma(H_qW_g),\\
 \widetilde O_h&=G_{:,h}\odot O_h,\\
 H_q'&=H_q+W_o[\widetilde O_1;\ldots;\widetilde O_H].
\end{aligned}
 \label{eq:evidence-gate}
\end{equation}
Here, $L$ is the number of decoder query positions, $H$ is the number of attention heads, $\sigma$ is the sigmoid function, and $\odot$ denotes element-wise multiplication with broadcasting over the head dimension. The learned matrix $W_g\in\mathbb R^{d\times H}$ produces $G\in(0,1)^{L\times H}$, which contains one gate per query and head; $O_h$ is the standard scaled dot-product output of head $h$, and $W_o$ projects the concatenated gated outputs. The gate allows each decoder layer to attenuate unreliable prefix updates before they enter the residual stream, preserving capacity for residue context supported by the spectrum.

\subsection{Autoregressive Semi-Supervision}
\label{sec:pseudo}
Following common semi-supervised practice~\cite{tarvainen2017mean,sohn2020fixmatch}, \method{} maintains an exponential moving average (EMA) teacher $\bar\theta$ and enforces consistency between weakly and strongly augmented spectrum views. For an unlabeled example $(x_u,c_u)\in\mathcal D_u$, the teacher generates pseudo-labels from a weak view $x_u^w$ that preserves canonical processing, while the student learns from a strong view $x_u^s$ with mild peak removal and intensity perturbation. Peak $m/z$ and precursor information remain unchanged across views. Because peptide tokens form an autoregressive and mass-dependent path, \method{} retains only a contiguous pseudo-prefix that is both mass-admissible and statistically reliable.

\textbf{Stepwise mass-constrained decoding.} Token-wise pseudo-labeling ignores a defining property of DNPS: every generated residue consumes part of a fixed precursor mass. Consequently, a locally confident but mass-incompatible token can corrupt all subsequent pseudo-labels. \method{} therefore constrains each autoregressive decision by the remaining precursor mass. Let $M_u=M(c_u)$ and let $m_{\min}$ be the lightest residue mass in $\mathcal V_{\mathrm{res}}$. Appending candidate residue $a$ to prefix $\hat y_{<t}$ leaves residual mass $r_t(a)=M_u-m(\hat y_{<t}a)$. The admissible vocabulary is
\begin{align}
 \mathcal V_t={}&\{a\in\mathcal V_{\mathrm{res}}:
 |r_t(a)|\leq\delta\ \lor\ r_t(a)\geq m_{\min}\}\notag\\
 &\cup\{\mathrm{EOS}:|M_u-m(\hat y_{<t})|\leq\delta\}.
 \label{eq:admissible-set}
\end{align}
Let $p_t(a)$ be the teacher's full-vocabulary probability for token $a$, conditioned on the current prefix, weak spectrum view, and precursor. We select the most probable mass-admissible token,
\begin{equation}
 \hat y_t=\arg\max_{a\in\mathcal V_t}p_t(a).
 \label{eq:ar-pseudo}
\end{equation}
The selected token retains its original full-vocabulary probability, ensuring that chemical masking does not alter the confidence used for pseudo-label selection.

\textbf{Path-consistent prefix selection.} The reliability of an autoregressive residue depends on every preceding decision. \method{} therefore evaluates the joint confidence of each contiguous generation path. With $q_t=p_t(\hat y_t)$ denoting the confidence of the selected residue conditioned on the preceding prefix and $\tau\in(0,1)$ the path-confidence threshold, we define
\begin{equation}
\begin{aligned}
 C_0&=1,\qquad C_t=C_{t-1}q_t=\prod_{i=1}^{t}q_i,\\
 T^*&=\max\bigl(\{0\}\cup\{t:C_t\geq\tau\}\bigr).
\end{aligned}
 \label{eq:confidence}
\end{equation}
The accepted target is the longest reliable prefix $\hat y_{1:T^*}$; if the first token fails, $T^*=0$ and the spectrum contributes no pseudo-label. Every retained residue is supported by its preceding path, while later suffix uncertainty does not remove reliable earlier supervision. Since $0\leq q_t\leq1$, $C_t$ is monotone non-increasing. Teacher decoding stops when $C_t<\tau$, at mass-valid EOS, when the admissible set is empty, or at the maximum pseudo-sequence length $T_{\max}$.

\textbf{Semi-supervised objective.} Given minibatches $B_\ell$ and $B_u$, the student observes $x_u^s$ and is teacher-forced on each accepted prefix. Let $T_u^*$ denote the accepted prefix length of example $u$. The loss is normalized by the number of accepted tokens,
\begin{equation}
 \mathcal L_{\mathrm{unsup}}=
 -\frac{\displaystyle\sum_{u\in B_u}\sum_{t=1}^{T_u^*}
 \log p_\theta(\hat y_{u,t}\mid\hat y_{u,<t},x_u^s,c_u)}
 {\displaystyle\sum_{u\in B_u}T_u^*}.
 \label{eq:unsup}
\end{equation}
The loss is set to zero when no token is accepted. The complete objective is $\mathcal L=\mathcal L_{\mathrm{sup}}+\lambda_u\mathcal L_{\mathrm{unsup}}$, where $\lambda_u$ controls the unlabeled term. Labeled spectra receive the same mild training perturbation as the strong unlabeled view, and labeled peptides and accepted pseudo-prefixes share one student forward pass. Algorithm~\ref{alg:seminovo} summarizes the training procedure.

\begin{algorithm}[t]
\caption{SemiNovo semi-supervised training}
\label{alg:seminovo}
\small
\begin{algorithmic}[1]
\Require batches $B_\ell,B_u$; student $\theta$; teacher $\bar\theta$; threshold $\tau$; weight $\lambda_u$; EMA decay $\beta$
\State Construct weak and strong views $(X_u^w,X_u^s)$ from $B_u$
\State Initialize accepted prefix set $\mathcal P\gets\varnothing$
\For{$u=1,\ldots,|B_u|$}
  \State $\hat y_u\gets\varnothing$ and $\ell_{C,u}\gets0$
  \For{$t=1,\ldots,T_{\max}$}
    \State Construct $\mathcal V_{u,t}$ using Eq.~\eqref{eq:admissible-set}
    \If{$\mathcal V_{u,t}=\varnothing$} \textbf{break} \EndIf
    \State Select $\hat y_{u,t}\gets\arg\max_{a\in\mathcal V_{u,t}}p_{u,t}(a)$
    \State $\ell_{C,u}\gets\ell_{C,u}+\log p_{u,t}(\hat y_{u,t})$
    \If{$\ell_{C,u}<\log\tau$} \textbf{break} \EndIf
    \State Append $\hat y_{u,t}$ to $\hat y_u$
    \If{$\hat y_{u,t}=\mathrm{EOS}$} \textbf{break} \EndIf
  \EndFor
  \If{$|\hat y_u|>0$} $\mathcal P\gets\mathcal P\cup\{(X_u^s,c_u,\hat y_u)\}$ \EndIf
\EndFor
\State Pack $B_\ell$ and $\mathcal P$ into one student forward pass
\State Update $\theta$ using $\mathcal L_{\mathrm{sup}}+\lambda_u\mathcal L_{\mathrm{unsup}}$
\State $\bar\theta\gets\beta\bar\theta+(1-\beta)\theta$
\end{algorithmic}
\end{algorithm}

\begin{table*}[!t]
  \caption{\DeNovo{} peptide sequencing performance under the NovoBench protocol. We report peptide precision and area under the precision--coverage curve (AUC) together with amino-acid precision and recall. Best and second-best results are shown in bold and underlined, respectively.}
  \label{tab:main}
  \centering
  \setlength{\tabcolsep}{1.5pt}
  \begin{tabular*}{\textwidth}{@{\extracolsep{\fill}}lcccccccccccc@{}}
    \toprule
    & \multicolumn{6}{c}{Peptide level} & \multicolumn{6}{c}{Amino-acid level} \\
    \cmidrule(lr){2-7}\cmidrule(lr){8-13}
    & \multicolumn{2}{c}{Nine-species} & \multicolumn{2}{c}{HC-PT} & \multicolumn{2}{c}{Seven-species}
    & \multicolumn{2}{c}{Nine-species} & \multicolumn{2}{c}{HC-PT} & \multicolumn{2}{c}{Seven-species} \\
    Method & Prec. & AUC & Prec. & AUC & Prec. & AUC & Prec. & Recall & Prec. & Recall & Prec. & Recall \\
    \midrule
    DeepNovo       & 0.428 & 0.376 & 0.313 & 0.255 & 0.204 & 0.136 & 0.696 & 0.638 & 0.531 & 0.534 & 0.492 & 0.433 \\
    PointNovo      & 0.480 & 0.436 & 0.419 & 0.373 & 0.022 & 0.007 & 0.740 & 0.671 & 0.623 & 0.622 & 0.196 & 0.169 \\
    Casanovo       & 0.481 & 0.439 & 0.211 & 0.177 & 0.119 & 0.084 & 0.697 & 0.696 & 0.442 & 0.453 & 0.322 & 0.327 \\
    AdaNovo        & 0.505 & 0.469 & 0.212 & 0.178 & 0.174 & 0.135 & 0.698 & 0.709 & 0.442 & 0.451 & 0.379 & 0.385 \\
    $\pi$-HelixNovo & 0.517 & 0.453 & 0.356 & 0.318 & 0.234 & 0.173 & 0.765 & 0.758 & 0.588 & 0.582 & 0.481 & 0.472 \\
    SearchNovo     & 0.550 & 0.489 & 0.447 & 0.413 & 0.259 & \underline{0.174} & 0.748 & 0.746 & 0.652 & 0.658 & \underline{0.489} & \underline{0.488} \\
    ReNovo         & 0.568 & 0.528 & 0.467 & 0.436 & \underline{0.278} & \textbf{0.228} & 0.770 & 0.769 & 0.651 & 0.648 & \textbf{0.512} & \textbf{0.514} \\
    DiffuNovo      & 0.572 & 0.413 & 0.485 & 0.324 & 0.233 & 0.104 & 0.785 & 0.783 & 0.648 & 0.648 & 0.430 & 0.428 \\
    \midrule
    \method{} (supervised)
      & \underline{0.582} & \underline{0.538}
      & \underline{0.490} & \underline{0.454}
      & 0.255 & 0.168
      & \textbf{0.800} & \underline{0.798}
      & \underline{0.665} & \underline{0.663}
      & 0.425 & 0.424 \\
    \method{}
      & \textbf{0.635} & \textbf{0.593}
      & \textbf{0.514} & \textbf{0.475}
      & \textbf{0.280} & 0.165
      & \underline{0.799} & \textbf{0.800}
      & \textbf{0.676} & \textbf{0.675}
      & 0.426 & 0.427 \\
    \bottomrule
  \end{tabular*}
\end{table*}

\textbf{Computational complexity.}
The encoder--decoder costs $\mathcal O(P^2d+T^2d+PTd+(P+T)d^2)$ per layer. Decoder gating adds only $\mathcal O(TdH)$, leaving the asymptotic inference complexity unchanged, while monotone early termination limits the training-only length of teacher decoding.

%% file: sections/04_experiments.tex
\section{Experiments}
\label{sec:experiments}

\begin{table}[t]
  \caption{PTM precision on the three NovoBench datasets.}
  \label{tab:ptm}
  \centering
  \small
  \setlength{\tabcolsep}{3pt}
  \begin{tabular*}{\columnwidth}{@{\extracolsep{\fill}}lccc@{}}
    \toprule
    Method & Nine-species $\uparrow$ & HC-PT $\uparrow$ & Seven-species $\uparrow$ \\
    \midrule
    DeepNovo & 0.576 & 0.626 & 0.391 \\
    PointNovo & 0.629 & 0.676 & 0.117 \\
    Casanovo & 0.706 & 0.501 & 0.360 \\
    AdaNovo & 0.652 & 0.552 & 0.448 \\
    $\pi$-HelixNovo & 0.680 & 0.568 & 0.473 \\
    SearchNovo & 0.764 & 0.715 & 0.472 \\
    DiffuNovo & 0.822 & 0.705 & \textbf{0.515} \\
    \midrule
    \method{} (supervised) & \underline{0.834} & \underline{0.731} & \underline{0.495} \\
    \method{} & \textbf{0.841} & \textbf{0.732} & 0.382 \\
    \bottomrule
  \end{tabular*}
\end{table}

\subsection{Experimental Settings}
\label{sec:setup}

\smallskip\noindent\textbf{Datasets.}
All experimental settings follow NovoBench~\cite{zhou2024novobench}. We evaluate on three datasets spanning different resolutions and biological origins: Nine-species~\cite{tran2017deepnovo}, HC-PT~\cite{eloff2023instanovo}, and Seven-species~\cite{tran2017deepnovo}. Evaluation uses the official NovoBench implementation of peptide precision, precision--coverage AUC, amino-acid precision and recall, and PTM precision. \dataset{} provides 4.5 million unlabeled spectra refined from 44,963,379 spectra across 43 public PRIDE projects. The same corpus and preprocessing are used for all targets without target-specific selection.

\smallskip\noindent\textbf{Implementation details.}
\method{} uses nine encoder and nine decoder blocks, hidden width 512, eight attention heads, and feed-forward width 1,024. We first train \method{} (supervised) from scratch for 30 epochs with batch size 32 and select the checkpoint with the highest validation sequence accuracy. Semi-supervised training starts from this checkpoint and jointly optimizes labeled NovoBench and unlabeled \dataset{} batches of 512 spectra for six unlabeled-data epochs, cycling the smaller labeled split. We use AdamW with peak learning rates of $5\times10^{-4}$ and $10^{-3}$ for the supervised and semi-supervised stages, EMA decay 0.999, $\lambda_u=1$, $\tau=0.9$, and maximum pseudo-sequence length 32.

\begin{figure*}[!t]
  \centering
  \includegraphics[width=\textwidth]{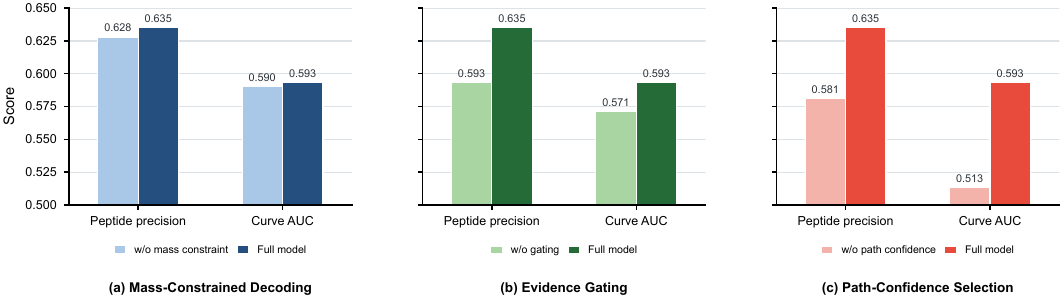}
  \caption{Three central ablations on Nine-species. Grouped bars compare the complete model with removal of stepwise mass constraints, removal of chemical-evidence gating, and replacement of cumulative path confidence by mean token confidence.}
  \label{fig:ablation}
  \Description{Three grouped-bar panels compare the full model with variants without mass constraints, chemical-evidence gating, or cumulative path confidence.}
\end{figure*}

\subsection{Main Results}
\label{sec:comparison}

\smallskip\noindent\textbf{Chemical-evidence gating yields a strong supervised model.}
Even without unlabeled spectra, \method{} (supervised) substantially improves over Casanovo on all three benchmarks. Peptide precision rises from 0.481 to 0.582 on Nine-species, from 0.211 to 0.490 on HC-PT, and from 0.119 to 0.255 on Seven-species. Curve AUC and amino-acid metrics improve consistently, including gains of more than twenty points in amino-acid precision and recall on HC-PT. Because this comparison uses only labeled NovoBench spectra, it establishes the gated spectrum-to-peptide model as a strong backbone before semi-supervised learning. The result is consistent with our motivation: suppressing chemically weak prefix updates allows diagnostic fragment evidence to contribute more effectively to sequence generation.

\smallskip\noindent\textbf{Unlabeled spectra primarily improve complete-peptide recovery.}
Adding \dataset{} raises peptide precision from 0.582 to 0.635 on Nine-species, from 0.490 to 0.514 on HC-PT, and from 0.255 to 0.280 on Seven-species. Nine-species and HC-PT achieve the strongest overall results, with AUC increasing to 0.593 and 0.475; Seven-species attains the highest peptide precision among the compared methods but not the best AUC or residue-level performance. Amino-acid metrics change only marginally on Nine-species and Seven-species and by about one point on HC-PT. Semi-supervision therefore does not simply make every residue easier to classify. Instead, exposure to unlabeled spectral contexts reduces the small number of local errors that prevent an otherwise nearly correct sequence from becoming an exact peptide match. Since peptide precision requires every residue to be correct, modest residue-level changes can produce substantially larger sequence-level gains.

\smallskip\noindent\textbf{All three benchmarks exhibit domain-dependent transfer.}
\dataset{} is assembled from heterogeneous acquisitions and is not identically distributed to any NovoBench benchmark. This shift appears differently across the three targets. On Nine-species, semi-supervision produces large peptide-level gains but almost no amino-acid or PTM gain, indicating that compatible high-resolution contexts improve sequence consistency without substantially changing local residue discrimination. HC-PT shows smaller peptide-level gains, one-point amino-acid gains, and nearly unchanged PTM precision, consistent with a moderate mismatch in peptide composition and cleavage regime. Seven-species exhibits the strongest shift: peptide precision improves, but curve AUC decreases slightly from 0.168 to 0.165 and PTM precision falls from 0.495 to 0.382. Thus, useful sequence patterns transfer across all three domains, whereas confidence calibration and modification-specific evidence are more sensitive to acquisition mismatch.

\smallskip\noindent\textbf{PTM identification reveals the same division of labor.}
Table~\ref{tab:ptm} shows that the supervised \method{} backbone raises PTM precision over Casanovo from 0.706 to 0.834 on Nine-species, from 0.501 to 0.731 on HC-PT, and from 0.360 to 0.495 on Seven-species. Chemical-evidence gating is therefore especially valuable for modified residues, whose localization depends on a small number of diagnostic fragment ions. Semi-supervision provides only modest additional gains on Nine-species and HC-PT, reaching 0.841 and 0.732, while Seven-species decreases to 0.382 despite stable amino-acid metrics. This disproportionate decline indicates confidence-calibration error under domain shift: confidence learned from \dataset{} is less aligned with modified-residue correctness in low-resolution Seven-species spectra, causing pseudo-label selection to provide less reliable PTM supervision even when ordinary residue accuracy remains stable.

\subsection{Analysis and Ablations}
\label{sec:ablation}

We conduct all analyses on Nine-species because it combines multiple biological species, modified residues, and high-resolution fragment measurements. Unless stated otherwise, each variant changes only the named factor and retains the same architecture, data split, optimization schedule, and decoding configuration.

\smallskip\noindent\textbf{Component ablations.}
Figure~\ref{fig:ablation} isolates three designs introduced by \method{}. Removing stepwise mass constraints only reduces peptide precision and curve AUC from 0.635/0.593 to 0.628/0.590. Its principal benefit is computational: pruning precursor-incompatible residues prevents decoding effort from being spent on paths that cannot produce a chemically feasible peptide. Removing chemical-evidence gating causes a larger decline to 0.593/0.571, supporting the attention analysis in Figure~\ref{fig:attention-analysis}: regulating chemically weak updates is important for preserving diagnostic fragment evidence. Replacing cumulative path confidence with mean-token confidence, a matched generic SSL control, causes the largest degradation to 0.581/0.513 despite validation tuning over $\{0.85,0.90,0.95\}$. A low-confidence erroneous residue can be diluted by confident predictions at other positions and enter the pseudo-label. Because subsequent residues condition on the erroneous prefix, this local mistake propagates through the autoregressive sequence and introduces substantially noisier supervision.

\begin{figure}[!t]
  \centering
  \includegraphics[width=\columnwidth]{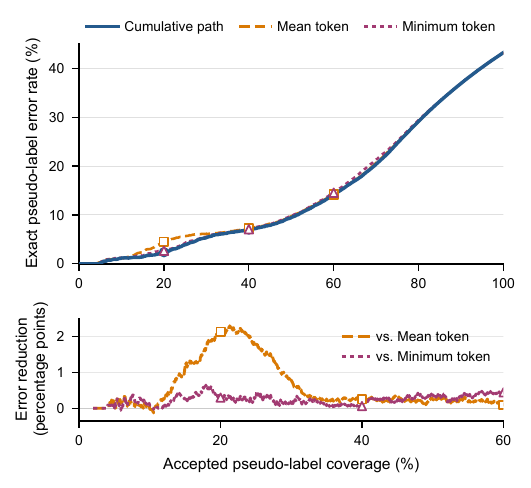}
  \caption{Exact pseudo-label error rate as a function of accepted coverage on the complete Nine-species validation split. The lower panel reports the error reduction of cumulative path confidence relative to the alternatives.}
  \label{fig:pseudo-reliability}
  \Description{Error--coverage curves compare cumulative path, mean token, and minimum token confidence, with a lower panel showing the error reduction from cumulative path confidence.}
\end{figure}

\smallskip\noindent\textbf{Pseudo-label reliability.}
We hide the annotations of all 28,572 Nine-species validation spectra during pseudo-label generation and reveal them only for evaluation. At matched coverage, cumulative path confidence yields the lowest exact-peptide error, with its largest advantage in the selective low-coverage region (Figure~\ref{fig:pseudo-reliability}). This result confirms that sequence-level path confidence ranks reliable autoregressive pseudo-labels more effectively than aggregating independent token scores.

\smallskip\noindent\textbf{Hyperparameter sensitivity.}
Table~\ref{tab:sensitivity} varies one factor at a time and reports validation performance. A path threshold of $\tau=0.90$ provides the strongest balance: lowering it to 0.85 introduces less reliable prefixes, whereas increasing it to 0.95 discards useful supervision. The unsupervised loss weight shows a similar interior optimum at $\lambda_u=1.0$; weaker weighting underuses dark spectra, while stronger weighting allows pseudo-label noise to dominate the labeled objective. Limiting pseudo-sequences to 32 residues also performs best, avoiding both premature truncation and unnecessary decoding of long, increasingly uncertain paths.

\smallskip\noindent\textbf{Effect of unlabeled-data scale.}
The final block of Table~\ref{tab:sensitivity} evaluates subsets of \dataset{}. Increasing the corpus from 0.5 million to 4.5 million spectra raises peptide precision from 0.580 to 0.631 and curve AUC from 0.523 to 0.595. Performance improves at each scale, establishing broad acquisition coverage as part of the data contribution. Quality filtering ensures valid model inputs, while near-duplicate control prevents recurring spectra from dominating this additional coverage.

\begin{table}[t]
  \caption{Hyperparameter and unlabeled-data sensitivity on the Nine-species validation split.}
  \label{tab:sensitivity}
  \centering
  \small
  \setlength{\tabcolsep}{3.0pt}
  \begin{tabular*}{\columnwidth}{@{\extracolsep{\fill}}lccc@{}}
    \toprule
    Factor & Value & Pep. prec. & Curve AUC \\
    \midrule
    \multirow{3}{*}{Path threshold $\tau$}
      & 0.85 & 0.621 & 0.572 \\
      & \textbf{0.90} & \textbf{0.631} & \textbf{0.595} \\
      & 0.95 & 0.628 & 0.588 \\
    \midrule
    \multirow{3}{*}{Loss weight $\lambda_u$}
      & 0.5 & 0.629 & 0.586 \\
      & \textbf{1.0} & \textbf{0.631} & \textbf{0.595} \\
      & 1.5 & 0.621 & 0.577 \\
    \midrule
    \multirow{3}{*}{Maximum pseudo length}
      & 24 & 0.629 & 0.586 \\
      & \textbf{32} & \textbf{0.631} & \textbf{0.595} \\
      & 40 & 0.630 & 0.590 \\
    \midrule
    \multirow{4}{*}{Unlabeled spectra}
      & 0.5M & 0.580 & 0.523 \\
      & 1.0M & 0.603 & 0.544 \\
      & 2.0M & 0.626 & 0.582 \\
      & \textbf{4.5M} & \textbf{0.631} & \textbf{0.595} \\
    \bottomrule
  \end{tabular*}
\end{table}

\begin{figure}[!t]
  \centering
  \includegraphics[width=\columnwidth]{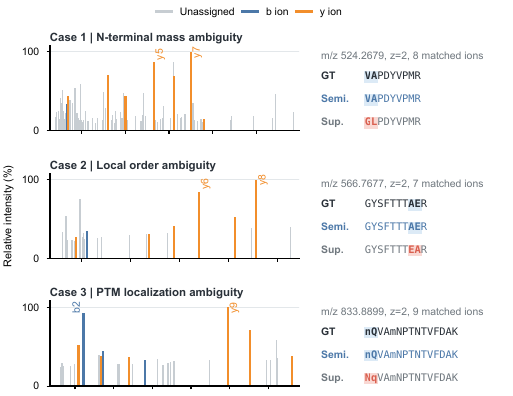}
  \caption{Three qualitative Nine-species cases. Each row places a compact tandem mass spectrum beside its sequence-level predictions. Gray sticks denote unassigned peaks; blue and orange sticks denote peaks matching theoretical b- and y-ions within 0.02 Da. Blue and red highlights mark the corresponding correct and erroneous sequence regions, respectively; lowercase residues denote their modified forms.}
  \label{fig:qualitative}
  \Description{Three annotated tandem mass spectra compare exact SemiNovo predictions with supervised predictions containing N-terminal, local-order, and modified-residue errors.}
\end{figure}

\smallskip\noindent\textbf{Qualitative analysis.}
Figure~\ref{fig:qualitative} presents three Nine-species spectra for which \method{} recovers the complete peptide while its supervised counterpart fails. The first prediction replaces the mass-equivalent N-terminal pair \texttt{VA} with \texttt{GL}; the second changes the local order \texttt{TAE} to \texttt{TEA} without altering peptide composition; and the third moves deamidation from \texttt{N} to the adjacent \texttt{Q}. These alternatives satisfy nearly identical precursor-mass constraints, but the highlighted b- and y-ion evidence supports the \method{} sequence. The cases show that unlabeled-spectrum learning improves local fragment interpretation where precursor feasibility alone cannot distinguish competing candidates.

%% file: sections/05_conclusion.tex
\section{Conclusion}
\label{sec:conclusion}
We study how search-conditioned supervision bias affects database-free \denovo{} inference. \method{} addresses this bias using \dataset{}, mass-constrained cumulative-path pseudo-labeling, and chemical-evidence gating. Together, these components extend supervision beyond identified spectra and improve the use of sparse fragment evidence. Unlabeled spectra improve peptide precision on all three benchmarks, while \method{} achieves state-of-the-art results on Nine-species and HC-PT. The remaining sensitivity on Seven-species highlights domain alignment as an important direction for learning from heterogeneous public spectra.

%% file: sections/06_limitations_ethics.tex
\section{Limitations and Ethical Considerations}
\label{sec:limitations}
\method{} formulates learning beyond search-identified spectra as a semi-supervised problem, but cross-dataset results also expose a domain-transfer challenge. \dataset{} is dominated by high-resolution acquisitions, and Seven-species reveals sensitivity to instrument resolution, fragmentation, and peptide distribution. Future work should combine semi-supervision with domain-aware corpus selection and acquisition conditioning.

The pseudo-label generator retains one mass-constrained path; preserving alternative feasible sequences may improve recall but requires efficient sequence-level calibration. Broader post-translational modification vocabularies also remain future work. All experiments use public PRIDE and NovoBench data and involve no new human-subject data collection.

%% file: sections/07_genai_usage.tex
\section{Generative AI Usage}
\label{sec:genai}
Generative AI tools assisted code development and manuscript drafting, translation, and editing. All AI-assisted material was reviewed by the authors, who retain responsibility for the research design, implementation, experiments, results, references, and final manuscript.

%% file: sections/06_appendix.tex
\appendix

\section{DarkSpec Corpus Details}
\label{app:darkspec}

\subsection{Source Scope}
\dataset{} is constructed from 653 acquisition runs in 43 public PRIDE projects. The source inventory contains 44,963,379 MS/MS spectra before model-compatible quality control. Table~\ref{tab:darkspec-sources} reports the contribution of every project; these counts are measured before filtering and redundancy control.

\begin{table}[H]
\caption{Public PRIDE projects in the \dataset{} source pool.}
\label{tab:darkspec-sources}
\centering
\small
\setlength{\tabcolsep}{2.7pt}
\begin{tabular}{lr@{\hspace{0.75em}}lr}
\toprule
Project & Spectra & Project & Spectra \\
\midrule
PXD055748 & 2,059,318 & PXD071883 & 251,927 \\
PXD060117 & 666,617 & PXD072146 & 933,697 \\
PXD060201 & 918,864 & PXD072226 & 467,463 \\
PXD060819 & 591,603 & PXD072402 & 217,986 \\
PXD061438 & 1,239,989 & PXD072579 & 88,232 \\
PXD061644 & 3,001,044 & PXD072877 & 983,211 \\
PXD061762 & 1,116,153 & PXD073076 & 5,228,025 \\
PXD061808 & 1,010,229 & PXD073471 & 427,952 \\
PXD061929 & 814,172 & PXD073702 & 1,536,478 \\
PXD062081 & 125,936 & PXD073864 & 177,875 \\
PXD069979 & 900,831 & PXD073985 & 3,959,311 \\
PXD070097 & 689,375 & PXD074179 & 3,934,079 \\
PXD070116 & 1,442,805 & PXD074303 & 258,239 \\
PXD070355 & 87,691 & PXD074339 & 226,303 \\
PXD070597 & 818,049 & PXD074388 & 1,604,360 \\
PXD070776 & 1,437,776 & PXD074405 & 647,384 \\
PXD071007 & 598,821 & PXD074406 & 523,248 \\
PXD071138 & 900,601 & PXD074452 & 378,687 \\
PXD071198 & 1,212,351 & PXD074558 & 346,772 \\
PXD071319 & 18,312 & PXD074667 & 431,124 \\
PXD071466 & 1,531,235 & PXD075330 & 189,170 \\
PXD071667 & 970,084 & & \\
\bottomrule
\end{tabular}
\end{table}

\subsection{Canonical Processing}
We apply the same peak representation to labeled NovoBench spectra and unlabeled PRIDE spectra. For each MS/MS spectrum, non-finite peaks and peaks outside $[50.52564895,2500]$~$m/z$ are removed, as are peaks within \SI{2}{\dalton} of the precursor. Intensities are first normalized by the base peak; peaks below 1\% relative intensity are discarded. We retain the 150 most intense remaining peaks, sort them by $m/z$, apply square-root intensity scaling, and renormalize by the base peak. Spectra are excluded when the precursor charge is outside $[1,10]$, peak arrays are malformed, or fewer than 20 valid peaks remain. Table~\ref{tab:darkspec-filtering} summarizes this reduction.

\begin{table}[H]
\caption{Quality control and experimental corpus size.}
\label{tab:darkspec-filtering}
\centering
\small
\setlength{\tabcolsep}{4pt}
\begin{tabular}{lr}
\toprule
Stage & Spectra \\
\midrule
Raw source pool & 44,963,379 \\
Rejected: fewer than 20 valid peaks & 3,700,995 \\
Rejected: invalid precursor charge & 166,308 \\
Rejected: malformed peak arrays & 7 \\
Quality-controlled pool & 41,096,069 \\
Fixed training corpus & 4,500,000 \\
\bottomrule
\end{tabular}
\end{table}

\subsection{Redundancy Control and Storage}
Pooling public projects can over-represent repeated exports and closely related acquisitions. We therefore assign each processed spectrum a near-duplicate key. Fragment $m/z$ values are rounded to integer bins and converted to a 32-bit SimHash using intensity-weighted random projections with a fixed seed. The key combines this spectral signature with precursor charge and precursor $\mathrm{MH}^{+}$ rounded to \SI{0.5}{\dalton}. One representative is retained per key. To bound repeated experimental cost, a fixed 4.5-million-spectrum subset of representatives is used for the reported training runs. The final corpus size is therefore a computational budget rather than an additional search-derived quality criterion.

Each retained example stores a $150\times2$ array of processed $m/z$--intensity pairs together with precursor $m/z$ and charge. No peptide sequence, database-search score, or search-derived confidence enters the unlabeled training records. The arrays are memory-mapped during training and converted to the same precursor and peak tensors used by the labeled NovoBench loader.

\subsection{Population Interpretation and Benchmark Overlap}
\dataset{} is acquisition-derived rather than search-failure-derived. Because membership does not depend on peptide assignment, it contains both spectra that could be identified by a suitable search and spectra that remain unidentified. \dataset{} consequently samples the acquisition-level spectrum population presented to a database-free model at inference. Instrument, sample, and acquisition differences remain visible, as quantified in Appendix~\ref{app:distribution}.

We audit data-level overlap between all 4.5 million \dataset{} spectra and every train, validation, and test split of the three NovoBench datasets. An exact match requires equality of the canonical $150\times2$ peak array, precursor $m/z$, and charge. We additionally flag conservative near-duplicate candidates using the same 32-bit SimHash, charge, and \SI{0.5}{\dalton} precursor-mass key used for \dataset{} redundancy control. Table~\ref{tab:dataset-overlap} shows no exact match in any split. Near-key candidates comprise 0.023--0.041\% of the test spectra, whereas semi-supervision improves peptide precision by 2.4--5.3 percentage points; their frequency is therefore insufficient to explain the measured gains.

\begin{table}[H]
\caption{Data-level overlap between \dataset{} and NovoBench. Near-key entries report candidate count and percentage of the benchmark split.}
\label{tab:dataset-overlap}
\centering
\small
\setlength{\tabcolsep}{2.8pt}
\begin{tabular}{llrrr}
\toprule
Dataset & Split & Spectra & Exact & Near-key (\%) \\
\midrule
\multirow{3}{*}{Nine-species}
 & Train & 499,402 & 0 & 220 (0.044) \\
 & Validation & 28,572 & 0 & 7 (0.025) \\
 & Test & 27,142 & 0 & 10 (0.037) \\
\midrule
\multirow{3}{*}{HC-PT}
 & Train & 213,284 & 0 & 58 (0.027) \\
 & Validation & 25,718 & 0 & 7 (0.027) \\
 & Test & 26,536 & 0 & 6 (0.023) \\
\midrule
\multirow{3}{*}{Seven-species}
 & Train & 317,009 & 0 & 77 (0.024) \\
 & Validation & 17,740 & 0 & 2 (0.011) \\
 & Test & 17,094 & 0 & 7 (0.041) \\
\bottomrule
\end{tabular}
\end{table}

\section{Implementation Details}
\label{app:implementation}

\subsection{Architecture and Optimization}
Table~\ref{tab:implementation} gives the configuration used for the reported experiments. The value of \texttt{n\_layers} is applied independently to the spectrum encoder and peptide decoder. Head-wise chemical-evidence gates are used in decoder causal self-attention. Isoleucine is normalized to leucine in target sequences, while oxidized methionine and deamidated asparagine and glutamine remain distinct vocabulary entries.

\begin{table}[H]
\caption{Model and optimization configuration.}
\label{tab:implementation}
\centering
\small
\setlength{\tabcolsep}{2.5pt}
\begin{tabular*}{\columnwidth}{@{}l@{\extracolsep{\fill}}l@{}}
\toprule
Component & Setting \\
\midrule
Encoder / decoder blocks & 9 / 9 \\
Hidden / feed-forward width & 512 / 1,024 \\
Attention heads & 8 \\
Trainable parameters & 48.2M \\
Dropout / weight decay & 0.1 / 0.05 \\
Maximum peaks / sequence length & 150 / 100 \\
Precursor tolerance / isotope errors & 50 ppm / $\{0,1\}$ \\
Optimizer / schedule & AdamW / warm-up and cosine decay \\
Numerical precision & bfloat16 mixed \\
Gradient clipping & 1.0 \\
Supervised epochs / batch / LR & 30 / 32 / $5\times10^{-4}$ \\
Semi-supervised epochs / batch pair & 6 / $512+512$ \\
Semi-supervised LR / warm-up & $10^{-3}$ / 2,000 steps \\
Teacher EMA decay / $\lambda_u$ & 0.999 / 1.0 \\
Path threshold / pseudo length & 0.9 / 32 \\
Peak dropout / intensity jitter & 0.05 / 0.02 \\
Random seed & 454 \\
\bottomrule
\end{tabular*}
\end{table}

The weak unlabeled view uses canonical processing without stochastic perturbation. The strong view independently drops each valid peak with probability 0.05 and multiplies retained intensities by factors sampled uniformly from $[0.98,1.02]$; precursor values and peak $m/z$ positions are unchanged. Supervised targets and accepted pseudo-prefixes use hard cross-entropy without label smoothing.

\subsection{Efficient Attention and Software Environment}
All spectrum-encoder self-attention, peptide-decoder causal self-attention, and encoder--decoder cross-attention layers use the official FlashAttention~2 kernels. Variable-length peak and peptide sequences are unpadded before attention, and fused dropout--residual--normalization kernels are used within each Transformer block. The proposed head-wise gate is applied only to decoder causal self-attention and multiplies each head output immediately before the standard output projection; the remaining attention computation is unchanged.

The reported experiments use Python~3.11, PyTorch~2.6, Lightning~2.1.4, CUDA~12, and FlashAttention~2.7.4.post1 on an NVIDIA H200 GPU. Optimization uses AdamW with default momentum coefficients $(0.9,0.999)$ and $\epsilon=10^{-8}$. The learning rate is updated after every optimization step using linear warm-up followed by cosine decay. Training uses bfloat16 mixed precision, gradient clipping at 1.0, and cuDNN benchmarking. Data loaders use 16 workers, pinned memory, persistent worker processes, and a prefetch factor of two.

\subsection{Epoch and Checkpoint Semantics}
Supervised training starts from random initialization and follows the NovoBench protocol for 30 epochs with batch size 32. The checkpoint with the highest validation sequence accuracy is reported as \method{}-Sup. and used to initialize semi-supervised training.

For the second stage, one epoch is defined by the larger of the labeled and unlabeled datasets. Let $N_\ell=|\mathcal D_\ell|$ and $N_u=|\mathcal D_u|$. A paired dataset has length $\max(N_\ell,N_u)$; examples from the smaller split are cycled while the paired indices are shuffled every epoch. Labeled and unlabeled minibatches each contain 512 spectra. With $N_u=4{,}500{,}000$, one epoch contains 8,789 optimization steps, and the reported six-epoch run contains 52,734 updates. This convention visits every unlabeled spectrum once per semi-supervised epoch while retaining labeled supervision at every update.

For confidence-criterion ablations, each criterion selects its best threshold from $\{0.85,0.90,0.95\}$ on the validation split.

Validation is performed every 1,000 updates. We retain only the checkpoint with the highest validation sequence accuracy and do not save a separate last checkpoint. Semi-supervised training uses an EMA teacher with decay 0.999; the online student is optimized, while teacher parameters are updated after each optimizer step.

\subsection{Open-source Resources}
The \method{} repository at \url{https://github.com/grandOrgan/Seminovo} provides the training and inference code, experiment configurations, \dataset{} construction pipeline, and NovoBench evaluation scripts. The versioned \dataset{} release is hosted on Hugging Face at \url{https://huggingface.co/datasets/PanLiu/DarkSpec}. Its Dataset Card documents the source PRIDE projects, data schema, quality-control procedure, provenance manifest, license, and loading interface.

\subsection{Mass-Constrained Pseudo-Label Implementation}
At each pseudo-decoding step, control tokens that cannot be emitted are first masked. Candidate residues are then screened against the precursor upper bound, and EOS is permitted only when the generated neutral peptide mass matches the observed precursor within 50 ppm under isotope errors 0 or 1 and the peptide contains at least six residues. The selected residue is the highest-probability member of this feasible set.

The probability used by cumulative path confidence is taken from the original softmax over the complete vocabulary rather than renormalized over the feasible set. Path confidence is accumulated in log space, and decoding terminates after confidence rejection, invalid mass progression, valid EOS, or 32 generated positions.

The student is teacher-forced only on the accepted contiguous prefix. Unsupervised cross-entropy is normalized by the total number of accepted pseudo-tokens in the minibatch rather than by the number of unlabeled spectra, so spectra with no reliable residue contribute zero loss. The supervised and accepted pseudo-labeled sequences are padded together and evaluated in one student forward pass.

\section{Evaluation Details}
\label{app:evaluation}

All reported metrics are computed with the official NovoBench sequence parser and mass-aware matcher. Let $N_{\mathrm{match}}^{\mathrm{aa}}$ be the number of matched residues, $N_{\mathrm{pred}}^{\mathrm{aa}}$ the number of predicted residues, and $N_{\mathrm{true}}^{\mathrm{aa}}$ the number of ground-truth residues. A residue match requires an individual mass difference below \SI{0.1}{\dalton} and a corresponding prefix or suffix cumulative-mass difference below \SI{0.5}{\dalton}. We report
\begin{equation}
 \operatorname{AA\text{-}Prec.}=\frac{N_{\mathrm{match}}^{\mathrm{aa}}}{N_{\mathrm{pred}}^{\mathrm{aa}}},
 \qquad
 \operatorname{AA\text{-}Rec.}=\frac{N_{\mathrm{match}}^{\mathrm{aa}}}{N_{\mathrm{true}}^{\mathrm{aa}}}.
\end{equation}
A peptide is correct only when every residue is matched under this rule. Peptide precision is the number of completely matched peptides divided by the number of test spectra, including spectra for which no sequence is returned. PTM precision is computed analogously over predicted modified-residue tokens. Curve AUC is obtained by sorting predictions by their model confidence and integrating the resulting peptide precision--coverage curve. The same evaluator, residue masses, modification vocabulary, and confidence aggregation are used for every compared method.

The ablation and threshold analyses use Nine-species with seed 454. Each component ablation changes only the named factor and otherwise retains the complete model configuration. For threshold analysis, validation spectra are treated as unlabeled when measuring pseudo-prefix precision, accepted-token coverage, and mean accepted-prefix length; test labels are never used for threshold selection.

\section{Supplementary Experiments}
\label{app:supplementary-experiments}

\subsection{Statistical Stability}
We repeat the complete Nine-species experiment with three random seeds while retaining the same data, optimization schedule, and model-selection criterion. Table~\ref{tab:seed-stability} reports a mean peptide precision of 0.632 and a mean curve AUC of 0.589. Across runs, the standard deviations are 0.009 and 0.005, respectively, and every run remains above the supervised checkpoint.

\begin{table}[H]
\caption{Statistical stability on Nine-species across three random seeds.}
\label{tab:seed-stability}
\centering
\small
\setlength{\tabcolsep}{5pt}
\begin{tabular}{lcc}
\toprule
Random seed & Pep. precision & Curve AUC \\
\midrule
454  & 0.635 & 0.593 \\
2,026 & 0.622 & 0.584 \\
1,145 & 0.639 & 0.591 \\
\midrule
Mean $\pm$ std. & $0.632\pm0.009$ & $0.589\pm0.005$ \\
\bottomrule
\end{tabular}
\end{table}

\subsection{Controlling for Additional Optimization}
Semi-supervised training introduces additional optimization steps after the supervised checkpoint. We isolate the contribution of unlabeled spectra by continuing supervised training for the same number of updates, with the same labeled batches, learning-rate schedule, and model-selection protocol. As shown in Table~\ref{tab:continuation-control}, continued supervised optimization yields only a small improvement, whereas adding \dataset{} produces a substantially larger gain. This result attributes the improvement to unlabeled-spectrum learning rather than training duration alone.

\begin{table}[H]
\caption{Matched-budget training control on Nine-species.}
\label{tab:continuation-control}
\centering
\small
\setlength{\tabcolsep}{4pt}
\begin{tabular}{lcc}
\toprule
Training setting & Pep. precision & Curve AUC \\
\midrule
Supervised checkpoint & 0.582 & 0.538 \\
Supervised continuation & 0.587 & 0.544 \\
\method{} & \textbf{0.635} & \textbf{0.593} \\
\bottomrule
\end{tabular}
\end{table}

\subsection{Pseudo-Label Reliability}
The matched generic SSL control retains the same teacher--student views, EMA, unlabeled data, optimization budget, and model initialization as \method{}, changing only cumulative path selection to mean-token confidence. At full coverage, exact-peptide pseudo-label accuracy on the 28,572 Nine-species validation spectra is 56.78\%. Table~\ref{tab:pseudo-reliability} complements the error--coverage curves in Figure~\ref{fig:pseudo-reliability} with accuracy at representative coverage levels.

\begin{table}[H]
\caption{Pseudo-label ranking quality on the Nine-species validation split.}
\label{tab:pseudo-reliability}
\centering
\small
\setlength{\tabcolsep}{2.6pt}
\begin{tabular}{lcccc}
\toprule
Confidence criterion & AUC & Acc.@20\% & Acc.@40\% & Acc.@60\% \\
\midrule
Cumulative path & \textbf{0.8531} & \textbf{97.65} & \textbf{92.99} & \textbf{85.89} \\
Mean token & 0.8493 & 95.50 & 92.75 & 85.80 \\
Minimum token & 0.8504 & 97.34 & 92.93 & 85.45 \\
\bottomrule
\end{tabular}
\end{table}

\subsection{Semi-Supervised Training Dynamics}
\begin{figure}[H]
\centering
\includegraphics[width=0.90\columnwidth]{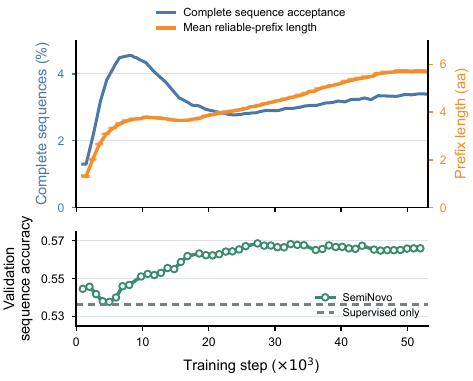}
\caption{Training dynamics for the Nine-species semi-supervised run initialized from the best supervised checkpoint. Reliable-prefix length, complete-sequence acceptance, and validation sequence accuracy follow distinct trajectories as the teacher adapts to 4.5 million unlabeled spectra.}
\label{fig:pseudo-dynamics}
\Description{Aligned training curves show a growing reliable-prefix length, non-monotonic complete-sequence acceptance, and validation sequence accuracy that peaks before slightly declining.}
\end{figure}

Figure~\ref{fig:pseudo-dynamics} tracks a Nine-species semi-supervised run initialized from the best supervised checkpoint. As the teacher adapts to unlabeled spectra, reliable prefixes become longer, while complete-sequence acceptance rises and then fluctuates because longer paths are more likely to contain a position that violates the cumulative confidence threshold. Validation sequence accuracy improves during the initial expansion of reliable pseudo-supervision, but declines after the midpoint even as pseudo-label coverage continues to grow. This divergence exposes a quantity--quality trade-off: increasing teacher confidence admits more and longer sequences, yet confidence growth does not eliminate domain-dependent pseudo-label errors, which are progressively reinforced through autoregressive training.

\subsection{Computational Efficiency}
We benchmark GPU-resident computation on a single NVIDIA H200 with a batch size of 512 for both training and inference. Throughput is computed from the median stable step time, and peak memory denotes the maximum allocated GPU memory. For semi-supervised training, labeled throughput counts the 512 labeled spectra in each update, whereas total throughput additionally counts the 512 unlabeled spectra processed by the teacher and student.

\begin{table}[H]
\caption{Single-H200 training efficiency at batch size 512.}
\label{tab:training-efficiency}
\centering
\small
\setlength{\tabcolsep}{2.0pt}
\begin{tabular}{lrrrr}
\toprule
Method & Params & Labeled/s & Total/s & Peak memory \\
\midrule
Casanovo & 47.35M & 3,513.9 & 3,513.9 & 14.60 GiB \\
\method{} (supervised) & 48.17M & \textbf{7,390.1} & \textbf{7,390.1} & 8.69 GiB \\
\method{} (semi-supervised) & 48.17M & 1,404.5 & 2,808.9 & 12.23 GiB \\
\bottomrule
\end{tabular}
\end{table}

\method{} (supervised) reaches 2.10$\times$ the training throughput of Casanovo while using 40.5\% less peak memory. Semi-supervised updates are slower because each step performs pseudo-label generation and optimization over both labeled and unlabeled spectra. The head-wise chemical-evidence gates introduce only 36,864 parameters, corresponding to 0.0765\% of the complete model.

\begin{table}[H]
\caption{Single-H200 inference efficiency at batch size 512.}
\label{tab:inference-efficiency}
\centering
\small
\setlength{\tabcolsep}{2.5pt}
\begin{tabular}{llrr}
\toprule
Method & Search & Spectra/s & Peak memory \\
\midrule
Casanovo & Original & 6.79 & 4.81 GiB \\
\method{} (supervised) & Casanovo-style & 93.15 & 3.04 GiB \\
\method{} (supervised) & Mass-pruned & \textbf{611.89} & 3.04 GiB \\
\method{} & Casanovo-style & 110.08 & 3.04 GiB \\
\method{} & Mass-pruned & \textbf{606.46} & 3.04 GiB \\
\bottomrule
\end{tabular}
\end{table}

Mass-feasibility pruning accelerates \method{} decoding by 6.57$\times$ for the supervised checkpoint and 5.51$\times$ for the complete model, without increasing peak memory. The optimized path processes approximately 89$\times$ more spectra per second than the original Casanovo implementation. The supervised and semi-supervised checkpoints share the same inference graph; pseudo-label generation introduces no deployment-time module.

\begin{table}[H]
\caption{Efficiency of pseudo-label search at batch size 512.}
\label{tab:pseudo-search-efficiency}
\centering
\small
\setlength{\tabcolsep}{2.8pt}
\begin{tabular}{lrrr}
\toprule
Pseudo-label search & Pseudo/s & Full step/s & Memory \\
\midrule
Fixed-length greedy & 1,516.0 & 2,309.4 & 12.89 GiB \\
Confidence early stopping & \textbf{2,187.7} & \textbf{3,092.0} & 12.24 GiB \\
Early stopping + mass pruning & 2,160.1 & 3,046.2 & 12.24 GiB \\
\bottomrule
\end{tabular}
\end{table}

Confidence-based early stopping provides the dominant pseudo-labeling speedup, improving search throughput by 44.3\% and complete-step throughput by 33.9\% over fixed-length greedy decoding. Adding mass-feasibility pruning retains nearly identical throughput, indicating that its primary role in pseudo-label generation is to prevent chemically infeasible continuations rather than to reduce the measured runtime.

\section{Dataset Distribution Analysis}
\label{app:distribution}

We compare \dataset{} with the labeled training splits of the three NovoBench datasets using 20,000 spectra sampled from each corpus. All statistics are computed after the canonical preprocessing in Appendix~\ref{app:darkspec}. Spectral entropy is defined as $-\sum_i p_i\log p_i$, where $p_i$ is the normalized intensity of peak $i$. Table~\ref{tab:dataset-distribution} reveals substantial acquisition-level differences. \dataset{} contains heavier precursors and a markedly larger proportion of charge-three-or-higher ions than any benchmark. Its retained-peak distribution is denser than Nine-species and HC-PT, while Seven-species exhibits the highest spectral entropy. These differences confirm that unlabeled training exposes the model to spectral conditions that are not represented by a simple resampling of any individual NovoBench dataset.

\begin{table}[H]
\caption{Distribution statistics after canonical preprocessing. Values are computed from 20,000 randomly sampled spectra per dataset.}
\label{tab:dataset-distribution}
\centering
\small
\setlength{\tabcolsep}{2.0pt}
\begin{tabular}{lrrrr}
\toprule
Dataset & Precursor $m/z$ & $z\geq3$ (\%) & Peaks & Entropy \\
\midrule
\dataset{} & 770.9 & 59.7 & 150 & 4.55 \\
Nine-species & 650.8 & 39.7 & 89 & 4.32 \\
HC-PT & 598.3 & 29.7 & 109 & 4.55 \\
Seven-species & 688.8 & 36.5 & 150 & 4.88 \\
\bottomrule
\end{tabular}
\end{table}